\documentclass[10pt,twocolumn,letterpaper]{article}

\usepackage{acpr}
\usepackage{times}
\usepackage{epsfig}
\usepackage{graphicx}
\usepackage{amsmath}
\usepackage{amssymb}
\usepackage{bm}
\usepackage{booktabs}

\usepackage[pagebackref=true,breaklinks=true,letterpaper=true,colorlinks,bookmarks=false]{hyperref}

\acprfinalcopy 

\ifacprfinal\fi
\begin{document}

\title{Irregular Convolutional Neural Networks}

\author{Jiabin Ma \quad Wei Wang \quad Liang Wang\\
Institution of Automation, Chinese Academy of Sciences\\
jiabin.ma@cripac.ia.ac.cn \quad wangwei@nlpr.ia.ac.cn \quad wangliang@nlpr.ia.ac.cn\\
}

\maketitle

\begin{abstract}
  Convolutional kernels are basic and vital components of deep Convolutional Neural Networks (CNN).
  In this paper, we equip convolutional kernels with shape attributes to generate the deep Irregular Convolutional Neural Networks (ICNN).
  Compared to traditional CNN applying regular convolutional kernels like ${3\times3}$, our approach trains irregular kernel shapes to better fit the geometric variations of input features.
  In other words, shapes are learnable parameters in addition to weights.
  The kernel shapes and weights are learned simultaneously during end-to-end training with the standard back-propagation algorithm.
  Experiments for semantic segmentation are implemented to validate the effectiveness of our proposed ICNN.
\end{abstract}

\section{Introduction}

In recent years, CNN becomes popular in both academia and industry as it plays a very important role in dealing with various feature extraction tasks. Practical and powerful as it is, there are still some problems that need to be discussed and solved.

Firstly, regular kernel shape in CNN mismatches with irregular feature pattern. One important fact in vision tasks is that though an input image has a fixed and regular size, the objects with irregular shapes are often of interest. Take image classification as an example, the target is to classify each object's category but not the whole image's category. Such situations are more obvious in detection and segmentation since the basic concept is to separate objects from the background. The feature pattern is general irregular. As the convolution operation is actually a dot product of two vectors, namely feature pattern and convolutional kernel, these two vectors should ideally have the same attributes to get more accurate response. In other words, the convolutional kernel should also be irregular as the input feature pattern does, so as to better extract most valuable information. But for traditional convolution, the kernel shape is often fixed, regular and can not be directly learned through the training process.

\begin{figure}[t]
\begin{center}
\begin{minipage}[t]{0.19\linewidth}
\end{minipage}
\hfill
\begin{minipage}[t]{0.3\linewidth}
  \centering
  \centerline{\includegraphics[height=0.8\linewidth,width=0.8\linewidth]{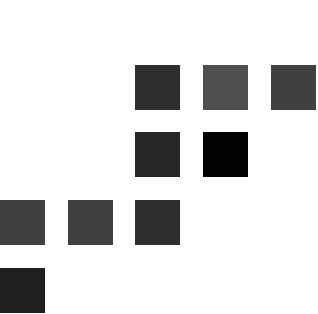}}
  \centerline{(a). Irregular input}\medskip
\end{minipage}
\hfill
\begin{minipage}[t]{0.3\linewidth}
  \centering
  \centerline{\includegraphics[height=0.85\linewidth,width=0.85\linewidth]{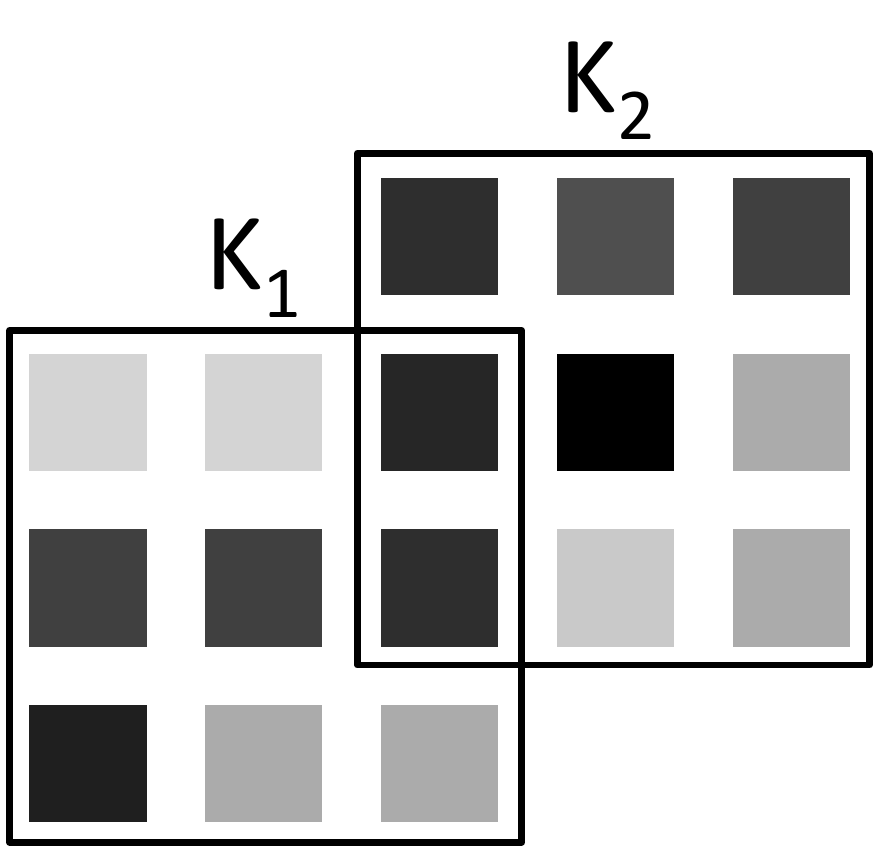}}
  \centerline{(b). Two ${3\times3}$ kernels}\medskip
\end{minipage}
\hfill
\begin{minipage}[t]{0.19\linewidth}
\end{minipage}
\vfill
\vspace{10pt}
\begin{minipage}[t]{1\linewidth}
  \centering
  \centerline{\includegraphics[height=0.23\linewidth,width=0.9\linewidth]{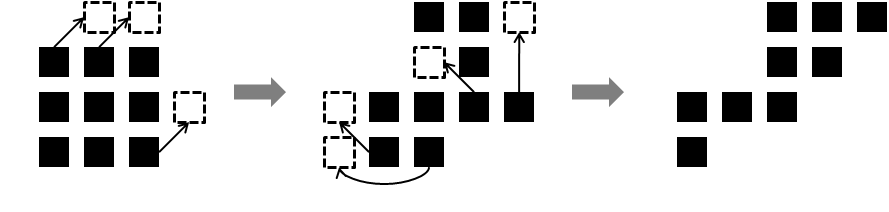}}
  \centerline{(c). Transformation from a ${3\times3}$ kernel to an irregular one}\medskip
\end{minipage}
\end{center}
   \caption{Comparison of regular and irregular convolutional kernels. (a) Irregular input excesses the receptive field of one ${3\times3}$ kernel. (b) ${K_1}$ and ${K_2}$ are two ${3\times3}$ kernels which are used together to model the input. (c) Shape transformation from a regular ${3\times3}$ kernel to an irregular one which fits the irregular input too. }
\label{fig:Figure1}
\end{figure}

Accordingly, the shape mismatch causes the ineffectiveness for regular convolutional kernel to model irregular feature pattern.
Convolutional kernel with a regular shape can also model irregular feature pattern, but it is not an effective way.
Its basic idea is that different scale weights distributed inside a regular shape can have similar effects as irregular ones.
As shown in Figure~\ref{fig:Figure1} (a) and (b), two convolutional kernels with ${3\times3}$ regular shape can model the irregular input pattern too.
But it is ineffective since it needs two kernels with 18 weights to model the input feature with 9 pixels.
It should be noted that things would get worse if the input feature has a longer or more discrete shape.
The inequality is caused by the fact that each of these two kernels needs many small weights, whose absolute values are close to zero, to obtain the irregular shape.
In other words, these small weights simply indicate the positions of big weights, but have no direct relationship with the input feature.
The phenomenon can also be illustrated in the task of model compression where weight pruning methods \cite{han2015deep, venkatesh2016accelerating} exactly cut the small kernel weights.

The above problems can be smoothly solved by using irregular convolutional kernel. Since regular kernel shape mismatches with irregular feature pattern, a most intuitive and reasonable method is to use irregular and trainable kernel shape. As illustrated in Figure~\ref{fig:Figure1} (c), a 9-weights irregular kernel has the same number of weights as a ${3\times3}$ regular kernel, but has similar function of two ${3\times3}$ regular kernels in which many weights are optimized to be zeros. So without stack, one 9-weights irregular kernel can also model a more complicated input feature, which explains the effectiveness in both function capacity and optimization. In conclusion, the key point of irregular convolutional kernel is to utilize trainable parameters and extract input features in a more effective way, which can help expand the network's modeling capacity with the same number of layers.

In order to apply irregular convolutional kernels on input features, we propose a new method to achieve transformation from a regular kernel shape to an irregular one. As shown in Figure~\ref{fig:Figure1} (c), the weights in a regular kernel would shift to new positions to find more valuable patterns, even the patterns are out of the original ${3\times3}$ receptive field. It is different from the pruning operation cutting bigger kernels, the training process can directly learn the shape variables, and there is no need to enlarge the kernel first. Considering the computation efficiency in practice, we reduce the number of shape variables and find that the addition of trainable variables does not aggravate the influence of overfitting. Experiments for semantic segmentation have been conducted to validate the effectiveness of our approach.

\section{Related Work}

In recent years, CNNs \cite{lecun1998gradient} greatly accelerate the development in computer vision tasks, including image classification \cite{lecun1998gradient,krizhevsky2012imagenet}, object detection \cite{girshick2015fast,ren2015faster,liu2016ssd} and semantic segmentation \cite{long2015fully,chen2016deeplab}. Deep stack of convolutional layers is able to model a sophisticated function, and Back Propagation \cite{hecht1988theory,lecun1990handwritten} makes it realizable to learn a large amount of parameters.

The series of Inception-VN \cite{szegedy2015going,ioffe2015batch,szegedy2016rethinking} make a lot of efforts to study kernel shapes. Inception V1 uses multi-scale kernels for scale-invariance, Inception V2 uses more small kernels to replace one bigger kernel, and Inception V3 uses the stack of one dimensional shape kernels to replace one square shape kernel. As we have mentioned above, though they have studied so many kinds of variants with respect to scale and shape, the kernel shape is still regular, fixed and not task optimized.

Effective Receptive Field \cite{luo2016understanding} helps us understand what input features are important and contribute a lot to the output of a network. It tells a truth that not all pixels inside theoretically receptive fields make expected contributions. This finding is coincident with our observation that there are many weights close to zero due to the incompatibility of regular kernel shape and irregular input shape.

Dilated convolution \cite{yu2015multi} expands distances between every two adjacent weights in a kernel, and generates a more discrete distribution. The presented dilated convolution is able to aggregate large scale contextual information without losing resolution. It is a variant of traditional compact kernels. In addition, our proposed ICNN is a further generalized version of dilated convolution.

Deformable Convolutional Networks (DCN) \cite{dai2017deformable} is a recently proposed excellent work which learns irregular kernels too. DCN has a similar thought with Region Proposal Network \cite{ren2015faster} as they both apply a normal convolution on the input feature and output the new receptive fields for the following operations with respect to each position in feature maps. There are two main insights in this work. First, kernel shapes vary from different spatial positions for the same input feature since that different parts of the same object often have different patterns. Second, kernel shapes vary from different inputs in the same spatial position so that the network could learn unique kernel shapes for different inputs. Different from DCN, we do not apply the above two insights but the shape variances across input channels as we think different feature maps have different feature patterns, in which dimension kernel shapes are remained same by default for DCN. Besides, since our approach is to insert shape attributes into the convolutional layer directly, it can simply update shape parameters by back propagation without any extra layers. We will show that our implementation of ICNN is more concise and can be easily applied to traditional convolutional layers.

\section{Irregular Convolutional Kernel}

\subsection{Basic Algorithm}

\textbf{\textit{Data structure:}} In our work, we add a new shift position attribute to the data structure of the convolutional kernel.
\begin{equation}
\begin{split}
& K=\left[ W,P \right] \\
& W=\left[ {{w}_{1}},{{w}_{2}},...,{{w}_{n}} \right] \\
& P=\left[ \left[ {{p}_{1,x}},{{p}_{1,y}} \right],\left[ {{p}_{2,x}},{{p}_{2,y}} \right],...,\left[ {{p}_{n,x}},{{p}_{n,y}} \right] \right] \\
\end{split}
\end{equation}
where ${W}$ is the set of original kernel weights. ${P}$ is the set of additional position attributes, and the origin of relative coordinate is set at the center of the regular kernel. ${n}$ is the number of weights.

\begin{figure}[t]
\begin{center}
\begin{minipage}[t]{0.49\linewidth}
  \centering
  \centerline{\includegraphics[height=0.9\linewidth,width=0.9\linewidth]{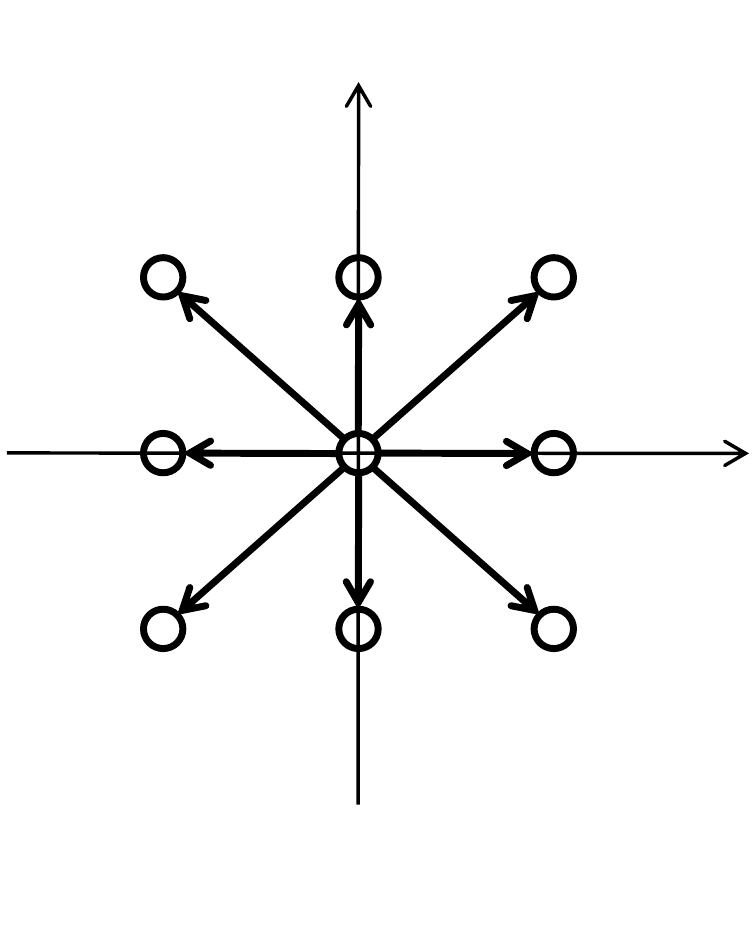}}
  \centerline{(a). Regular kernel}\medskip
\end{minipage}
\hfill
\begin{minipage}[t]{0.49\linewidth}
  \centering
  \centerline{\includegraphics[height=0.9\linewidth,width=0.9\linewidth]{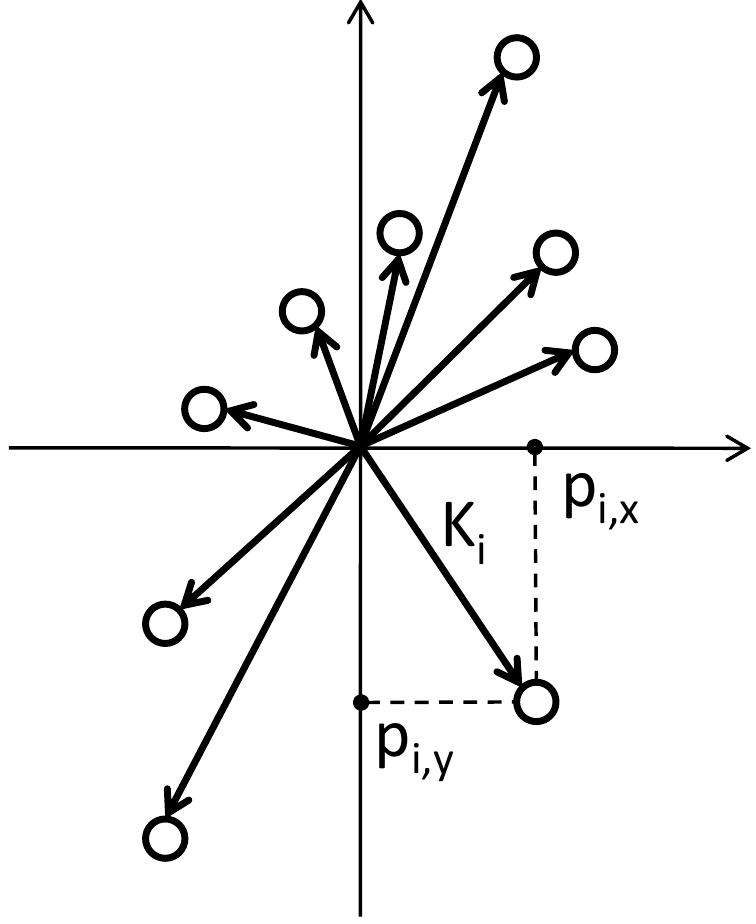}}
  \centerline{(b). Irregular kernel}\medskip
\end{minipage}
\vfill
\vspace{10pt}
\begin{minipage}[t]{0.6\linewidth}
  \centering
  \centerline{\includegraphics[height=0.9\linewidth,width=0.9\linewidth]{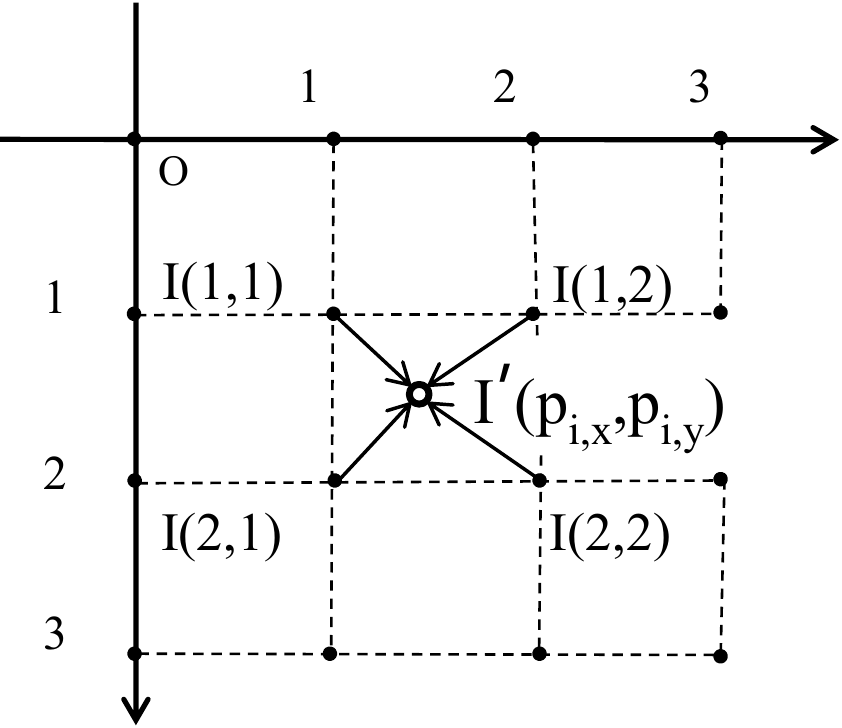}}
  \centerline{(c). Interpolation}\medskip
\end{minipage}
\end{center}
\caption{(a) Positions of a regular kernel are fixed in a square shape all time. (b) Positions of an irregular kernel are changing during training as gradients from loss function backward propagate to them. (c) Bilinear interpolation for float position ${[p_{i,x},p_{i,y}]}$. }
\label{fig:Figure2}
\end{figure}

\textbf{\textit{Shape interpolation:}} The convolutional output can be calculated as
\begin{equation}
OUT = \sum_{i} \sum_{j} w_i I_j \cdot 1(p_{i}=p_j^I)
\end{equation}
where ${I}$ means original input feature, and ${1(p_{i}=p_j^I)}$ is true only when the kernel's $\bm{i}$-th position ${p_{i}}$ equals to the input's $\bm{j}$-th position ${p_j^I}$.
The problem is that since ${p_j^I}$ is discrete, positions become non-differentiable.
To solve this problem, we change the discrete positions into continuous ones which are differentiable, \ie positions ${[p_{i,x},p_{i,y}]}$ should all be float numbers.
And bilinear interpolation is applied to obtain the continuous feature space.
For the $\bm{i}$-th interpolated input $I'_i$ whose position in the kernel's coordinate is ${[p_{i,x},p_{i,y}]}$
\begin{equation}
\begin{split}
\label{equation interp}
I'_i &=  f_{interp}(I) \\
&= \sum_{[x,y]}(1-|p_{i,x}- x|)(1-|p_{i,y}-y|)I(x,y) \\
&s.t. [x,y]\in N(p_{i,x},p_{i,y})
\end{split}
\end{equation}
where ${N(p_{i,x},p_{i,y})}$ means integer positions adjacent to ${[p_{i,x},p_{i,y}]}$, here are all 4 such positions in the spatial space.

\textbf{\textit{Convolution:}} Convolution is same as before except the replacement from ${I}$ to ${I'}$. For each output
\begin{equation}
\begin{split}
&OUT(r^o,c^o) = \sum_{i=1}^n w_i I'_i
\end{split}
\end{equation}
where ${[r^o,c^o]}$ means an absolute coordinate whose origin is set in the left-up corner of the output feature map, which is different from the relative coordinate of convolutional kernels.

\textbf{\textit{Initialization:}} The weights ${W}$ can be initialized from either random distribution or a trained model. Positions ${P}$ should be initialized by a regular shape, \ie, mostly used ${3\times3}$ squared shapes. But since the absolute function in bilinear interpolation is non-differentiable when ${[p_{i,x},p_{i,y}]}$ equal to integers, we should add a small shift to avoid this effect.
\begin{equation}
\begin{split}
&[p_{i,x},p_{i,y}] = [p_{i,x}^I + \epsilon, p_{i,y}^I + \epsilon]
\end{split}
\end{equation}
where ${[p_{i,x}^I, p_{i,y}^I]}$ mean regular integer positions as before, \eg, ${[-1,-1],[-1,0],[1,-1],...,[1,1]}$ for regular ${3\times3}$ kernel when the origin coordinate is set at the kernel center. And ${\epsilon}$ means a small shift added to the positions.

\textbf{\textit{Back propagation:}} Derivatives for ${W}$ and ${I'}$ are identical to the ones of traditional convolution except the replacement from ${I}$ to ${I'}$. The key problem is to compute the derivatives for ${I}$ and ${[p_{i,x},p_{i,y}]}$. Note that besides ${[p_{i,x},p_{i,y}]}$, all coordinates in back propagation are absolute coordinates.
Derivatives for ${I}$, ${p_{i,x}}$ and ${p_{i,y}}$ are denoted as
\begin{equation}
\begin{split}
&\Delta I({r},{c})=\sum\limits_{i=1}^{n}{\sum\limits_{{r}^{I},{c}^{I}}}D^r D^c\Delta I'({r^I},{c^I}) \\
& \Delta {{p}_{i,x}}=\sum\limits_{{{r}^{or}},{{c}^{or}}}\Delta I'({r^I},{c^I}) \cdot {\sum\limits_{r,c}{{{\left( -1 \right)}^{{{r}^{I}}>r}}D^c I\left( r,c \right)}} \\
& \Delta {{p}_{i,y}}=\sum\limits_{{{r}^{or}},{{c}^{or}}}\Delta I'({r^I},{c^I}) \cdot {\sum\limits_{r,c}{{{\left( -1 \right)}^{{{c}^{I}}>c}}D^r I\left( r,c \right)}} \\
&s.t.\text{ }\left\{ \begin{array}{ll}
 & D^r = \left( 1-\left| {r}-{{r}^{I}} \right| \right) \\
 & D^c = \left( 1-\left| {c}-{{c}^{I}} \right| \right) \\
 & {{r}^{I}}={{r}^{or}}+{{p}^{i,x}} \\
 & {{c}^{I}}={{c}^{or}}+{{p}^{i,y}} \\
 & {{r}^{or}}=floor({{r}}-{{p}_{i,x}})\text{  or  }ceil({{r}}-{{p}_{i,x}}) \\
 & {{c}^{or}}=floor({{c}}-{{p}_{i,y}})\text{  or  }ceil({{c}}-{{p}_{i,y}}) \\
 & 0<{{r}^{or}}<R^{I} \\
 & 0<{{c}^{or}}<C^{I} \\
 & {{r}^{or}}=k\times s^r \\
 & {{c}^{or}}=k\times s^c
\end{array} \right.
\end{split}
\end{equation}
For ${\Delta I(r,c)}$, first summation means derivatives come from all kernel weights, and second summation means each weight ${w_i}$ contributes to the input ${I}$'s derivatives multiple times so far as ${I}$ is adjacent to ${I'}$.
For ${\Delta p_{i,x}}$ and ${\Delta p_{i,y}}$, first summation means derivatives accumulate from all spatial positions when convolutional kernel is sliding, and second summation is the back propagation process of interpolation Eq.(\ref{equation interp}).
${[r,c]}$ is the integer position for ${I}$. ${[r^I,c^I]}$ is the float position for ${I'}$. ${[r,c]}$ and ${[r^I,c^I]}$ are adjacent to each other in every formulation. ${[r^{or},c^{or}]}$ is the integer position of the relative coordinate origin. ${[s^r, s^c]}$ is the convolutional stride and ${k}$ is of any integer. ${[R^{I},C^{I}]}$ means the spatial size of the input feature.

\subsection{Implementation}

\begin{figure}[t]
\begin{center}
\begin{minipage}[t]{0.49\linewidth}
  \centering
  \centerline{\includegraphics[height=0.9\linewidth,width=1.8\linewidth]{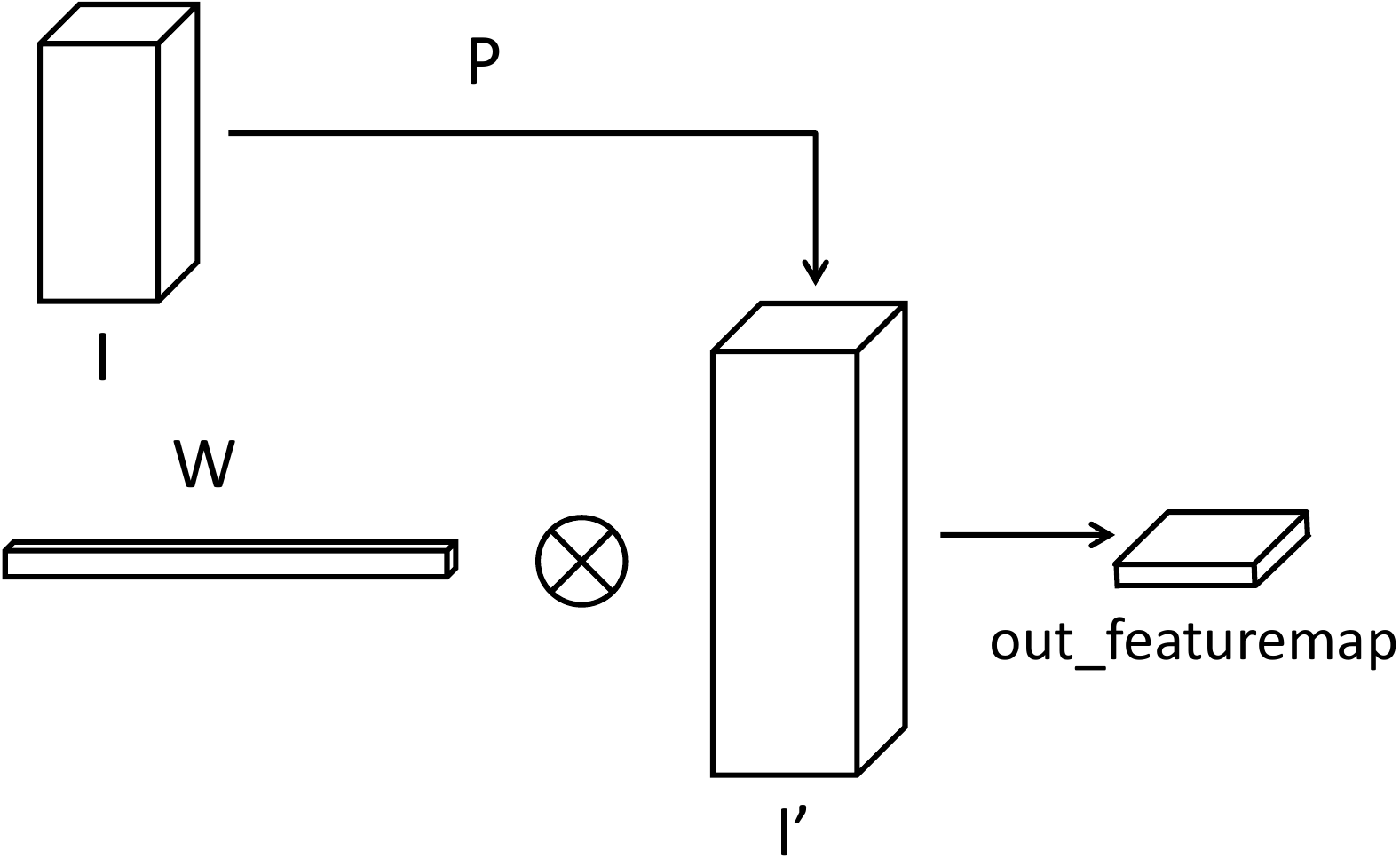}}
\end{minipage}
\end{center}
\caption{ Illustration for applying kernel's position attribute ${P}$ to input image and completing matrix multiplication with ${W}$.}
\label{fig:Figure3}
\end{figure}

In practice, the irregular position attribute ${P}$ is applied on input features. Same as ${im2col}$ which can accelerate convolution by matrix calculation and parallel computation, equipping the kernel with an irregular shape is equivalent to applying the rearrangement on the input. The key point is that the number of rearrangements equals to the number of different kernel shapes. For regular convolutional layer with ${N}$ output channels, as all ${N}$ kernels share the same shape \eg, ${3\times3}$, there is only one kind of rearrangement. But for irregular convolution, as each kernel has a unique shape, we should get ${N}$ different rearrangements with respect to ${N}$ different kinds of ${P}$, which rapidly increases computation. To solve this problem, we make proper adjustments. We let all the ${N}$ kernels share the same irregular shape. Whereas 1x1 kernels are not modified into irregular kernels since they are not feature extraction or combination functions but simply dimension transformations.

\section{Experiments}

\subsection{Result of Semantic Segmentation}

\textbf{\textit{Dataset:}} We validate our method on three widely used semantic segmentation datasets: PASCAL VOC 2012 \cite{everingham2015pascal}, PASCAL CONTEXT \cite{mottaghi2014role}, CITYSCAPE \cite{cordts2016cityscapes}.

PASCAL VOC 2012 segmentation dataset contains 20 object classes and 1 background class. The original dataset has 1,464 images for training, 1,449 images for validation and 1,456 images for test, respectively. And in \cite{hariharan2011semantic}, the annotations are augmented, leading to 10,582 images for training. Following the procedure of \cite{long2015fully}, we use mean intersection-over-union (mIoU) to evaluate the final performance. Base learning rates for kernel weights ${W}$ and positions ${P}$ are 0.001 and 50.0 respectively. We use the poly strategy for both learning rates. Max iteration is set 20K with a batch size of 16. In training phase, we use a crop size of 321 and basic data augmentations for scale and mirror variances.

PASCAL CONTEXT is a better annotation version of PASCAL VOC 2010, it is challenging as it has 60 categories. It has 4,998 images for training and 5,105 images for validation. Learning policy and basic preprocessing are same as the ones of PASCAL VOC 2012.

CITYSCAPE is a segmentation dataset focusing on city street scenes. It has 19 categories comparative to PASCAL VOC 2012. Every image in this dataset has a high resolution of ${2048\times1024}$, which makes a very hard problem to train a deep neural network in a memory limited GPU. Crop sizes are 473 for training and 713 for validation respectively. Max iteration is set to 45K.

\textbf{\textit{Model:}} The experiments are implemented with the most frequently-used deep residual network ResNet-101 \cite{he2016deep}, which has been pretrained on ImageNet \cite{deng2009imagenet}.

\begin{table}
\begin{center}
\begin{tabular}{ll}
    \toprule
    Layers with irregular kernels     & mIoU(\%) \\
    \midrule
    none (deeplab largeFOV \cite{chen2016deeplab}) & 72.8\\
    classifier     & 73.2\\
    from res5a to classifier     & 73.9\\
    from res4b11 to classifer      & 75.3\\
    from res4a to classifer    & 74.5\\
    \bottomrule
\end{tabular}
\end{center}
\caption{Ablation results on PASCAL VOC 2012 \cite{everingham2015pascal} validation dataset}
\end{table}

\begin{table}
\begin{center}
\begin{tabular}{ll}
    \toprule
    Methods     & mIoU(\%)\\
    \midrule
    deeplab largeFOV \cite{chen2016deeplab} & 41.7\\
    ICNN      & 43.1\\
    \bottomrule
\end{tabular}
\end{center}
\caption{Results on PASCAL CONTEXT \cite{mottaghi2014role} validation dataset}
\end{table}

\begin{table}
\begin{center}
\begin{tabular}{ll}
    \toprule
    Methods     & mIoU(\%)\\
    \midrule
    deeplab largeFOV \cite{chen2016deeplab} & 71.0\\
    ICNN      & 72.9\\
    \bottomrule
\end{tabular}
\end{center}
\caption{Results on CITYSCAPE \cite{cordts2016cityscapes} validation dataset}
\end{table}

\textbf{\textit{Update value:}} The updated position attribute ${P_{update}}$ should not excess the last value ${P_{last}}$ too much because the calculated interpolations and gradients are computed from the last adjacent positions ${P_{last}}$. In practice, we let
\begin{equation}
\begin{split}
floor(P_{last}) - \epsilon < P_{update} < ceil(P_{last}) + \epsilon
\end{split}
\end{equation}
where ${\epsilon}$ is a small positive value to allow ${P_{update}}$ get out of the last adjacent position ${P_{last}}$ but not too much.

\textbf{\textit{Result:}} Ablation studies are performed on PASCAL VOC 2012 and we take deeplab largeFOV \cite{chen2016deeplab} as the baseline. We can see that performance gets better with more irregular convolutional layers and saturates with applications of roughly half of the network, and finally gets 75.3\% mIoU. Using irregular kernels from the layer ${res4b11}$ same as experiment on PASCAL VOC 2012, we also test ICNN on PASCAL CONTEXT and CITYSCAPE, and finally gets 43.1\% and 72.9\% mIoUs respectively.

\subsection{Analysis of Kernel Shape}

\begin{figure}[t]
\begin{center}
\begin{minipage}[t]{0.49\linewidth}
  \centering
  \centerline{\includegraphics[height=0.95\linewidth,width=0.95\linewidth]{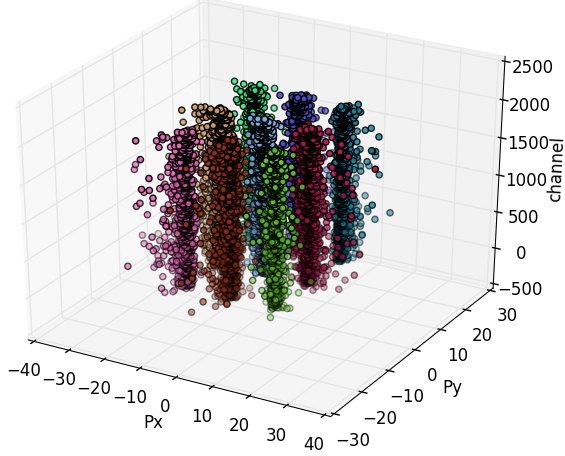}}
  \centerline{(a). fc1\_voc12 3D}\medskip
\end{minipage}
\hfill
\begin{minipage}[t]{0.49\linewidth}
  \centering
  \centerline{\includegraphics[height=0.95\linewidth,width=0.95\linewidth]{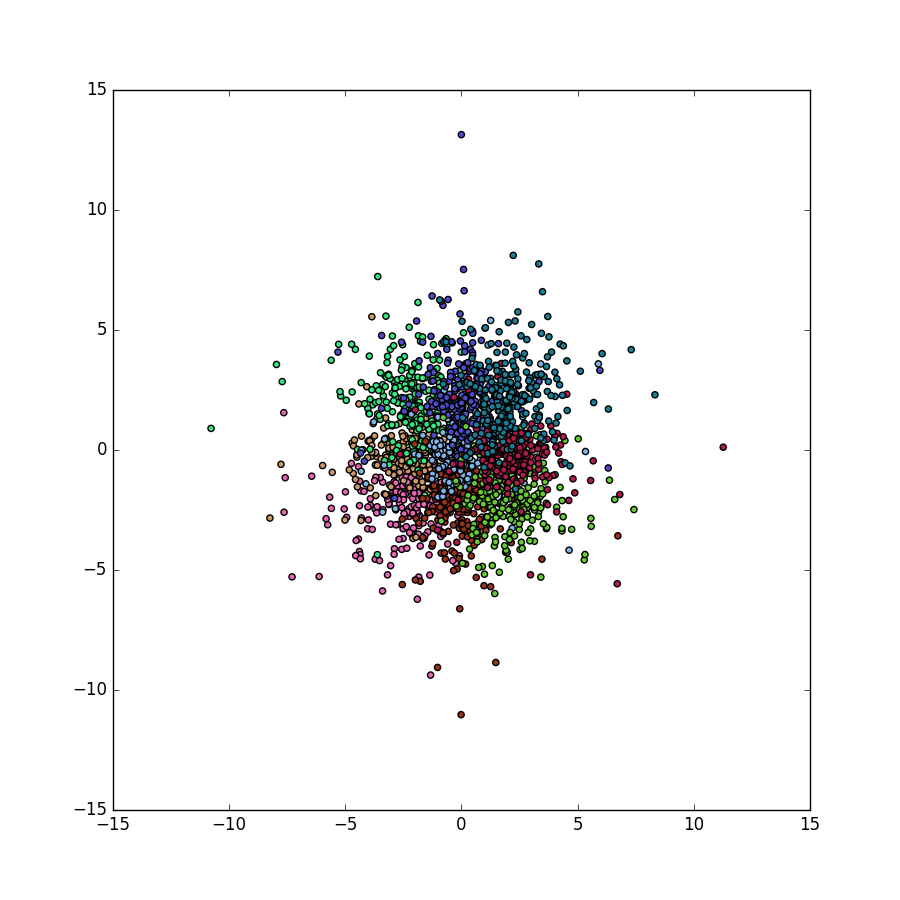}}
  \centerline{(b). res4b11}\medskip
\end{minipage}
\vfill
\begin{minipage}[t]{0.49\linewidth}
  \centering
  \centerline{\includegraphics[height=0.95\linewidth,width=0.95\linewidth]{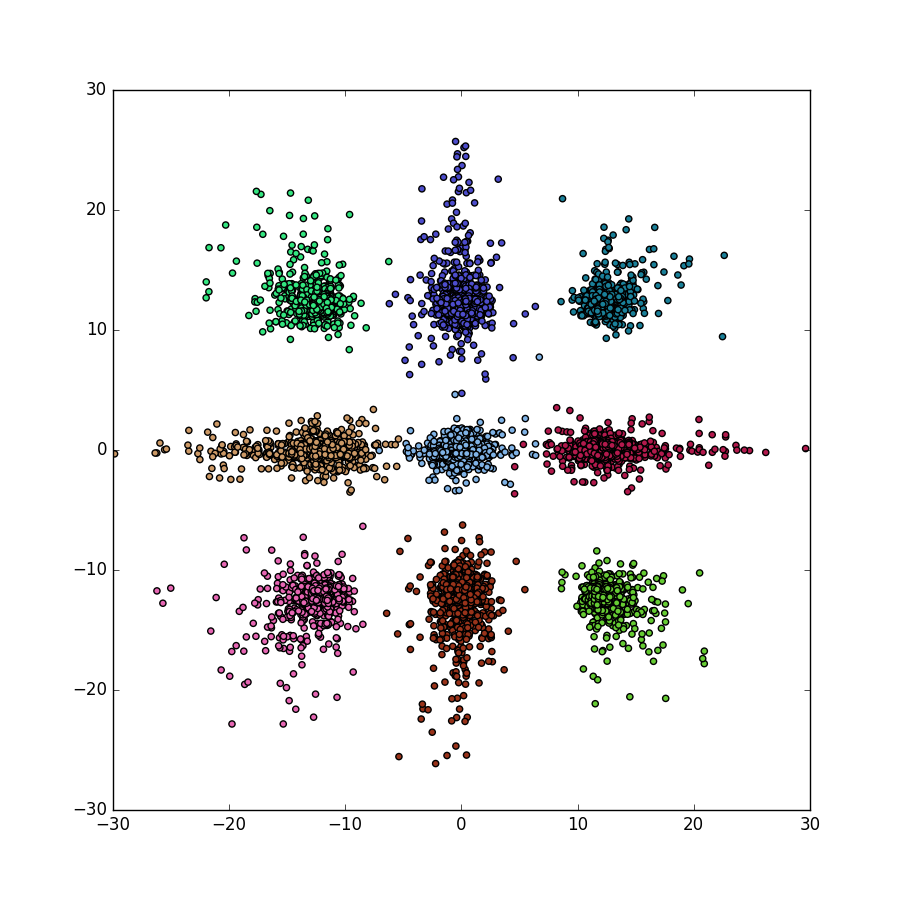}}
  \centerline{(c). fc1\_voc12}\medskip
\end{minipage}
\hfill
\begin{minipage}[t]{0.49\linewidth}
  \centering
  \centerline{\includegraphics[height=0.95\linewidth,width=0.95\linewidth]{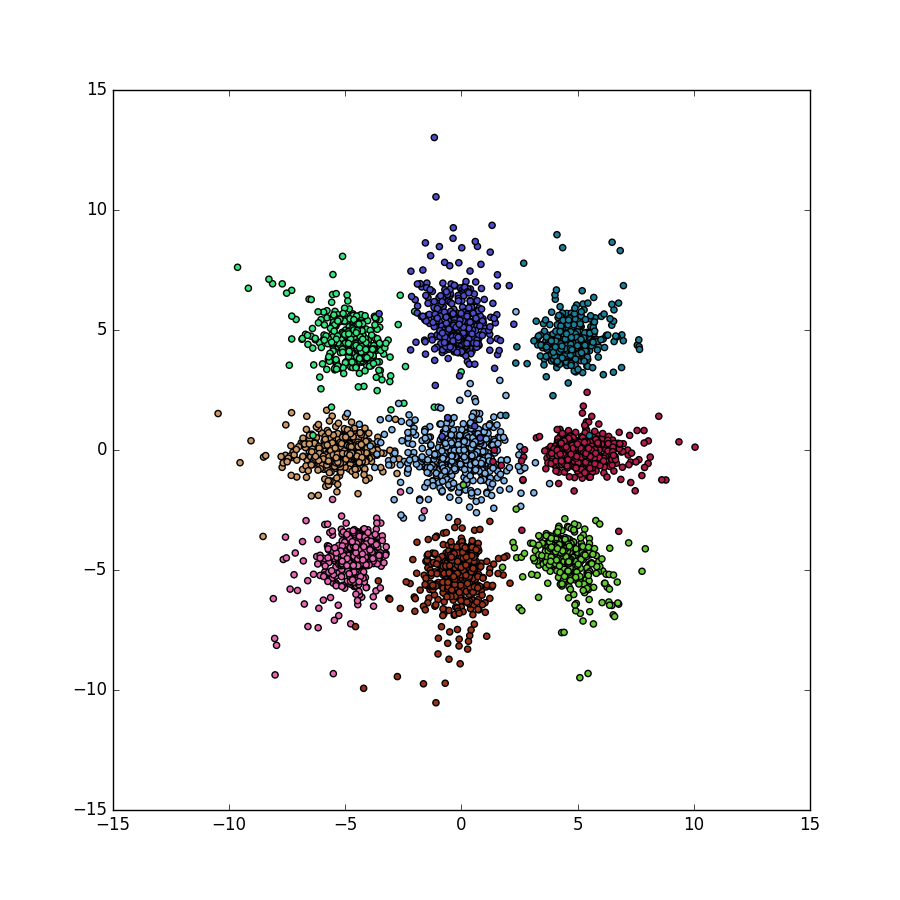}}
  \centerline{(d). res5c}\medskip
\end{minipage}
\end{center}
\caption{ Illustration of different kernel shapes for different layers respectively. (a) is the three dimensional visualization of the kernel shape for the last convolutional layer ${fc1\_voc12}$ and (c) is its projection on height-width in two dimensional space. (b) and (d) are both two dimensional projections for the corresponding layers. In these plots, pixels with same color mean they originally belong to same positions in the kernel. Best viewed in color.}
\label{fig:Figure4}
\end{figure}

In order to understand what exactly happens in kernel shape transformation, we visualize several kernels belonging to different layers in the network. The model is trained on the PASCAL VOC 2012 dataset.

\textbf{\textit{Comparison with original regular kernel:}} irregular kernel weights in one layer pay attention to variant scales and shapes instead of original fixed scale and square shape. We can find that weights which originally belong to the same positions have a rough Gaussian distribution, which have the same color as shown in Figure~\ref{fig:Figure4}. The centers of nine distributions roughly equal to original regular positions, but the variances of distributions respectively determine the abilities for extracting both global and detailed information as far as the training process asks them to be.

\textbf{\textit{Comparison among different layers:}} the comparison of Figure~\ref{fig:Figure4} (c) and others shows that the deeper layers' distributions are longer and more narrow. For ${fc1\_voc12}$, the surrounding eight distributions are more likely to have strip shapes, which can help deeper layers have a larger global receptive field with a limited number of weights.

\subsection{Heatmap Visualization for a Certain Pixel}

For a certain pixel in semantic segmentation, a proper receptive field is important for its inference category. We visualize heatmap with two steps as \cite{zeiler2014visualizing} does. We first apply the forward propagation. For the second step in back propagation, we choose a certain pixel, set all gradients to zeros except for the ones from the chosen pixel, back propagate the gradients to the input and get the gradient map as heatmap.

\begin{figure}[t]
\begin{center}
\begin{minipage}[t]{0.24\linewidth}
  \centering
  \centerline{\includegraphics[height=0.95\linewidth,width=0.95\linewidth]{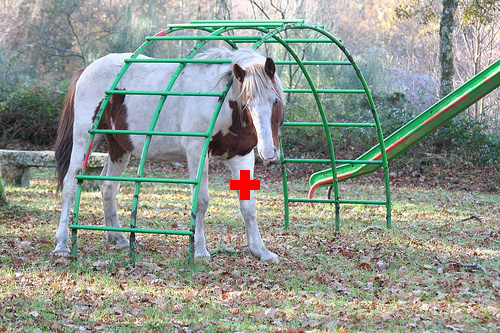}}
\end{minipage}
\hfill
\begin{minipage}[t]{0.24\linewidth}
  \centering
  \centerline{\includegraphics[height=0.95\linewidth,width=0.95\linewidth]{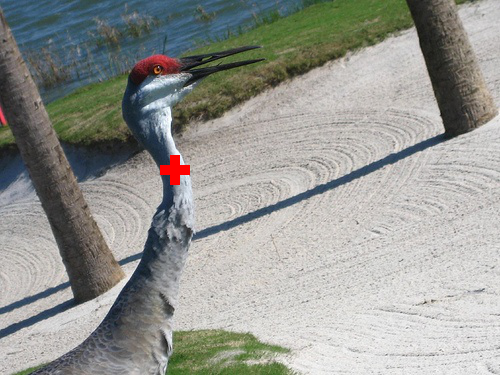}}
\end{minipage}
\hfill
\begin{minipage}[t]{0.24\linewidth}
  \centering
  \centerline{\includegraphics[height=0.95\linewidth,width=0.95\linewidth]{FOV_2/house2_point.png}}
\end{minipage}
\hfill
\begin{minipage}[t]{0.24\linewidth}
  \centering
  \centerline{\includegraphics[height=0.95\linewidth,width=0.95\linewidth]{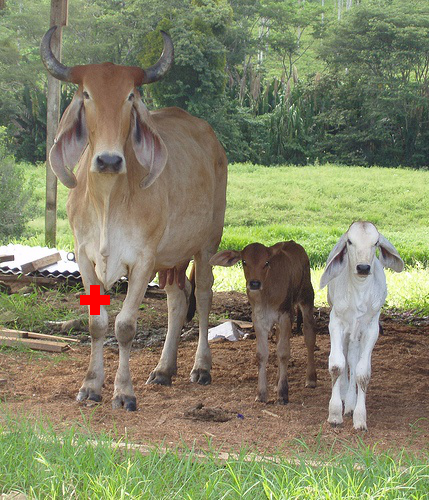}}
\end{minipage}
\vfill
\vspace{4pt}
\begin{minipage}[t]{0.24\linewidth}
  \centering
  \centerline{\includegraphics[height=0.95\linewidth,width=0.95\linewidth]{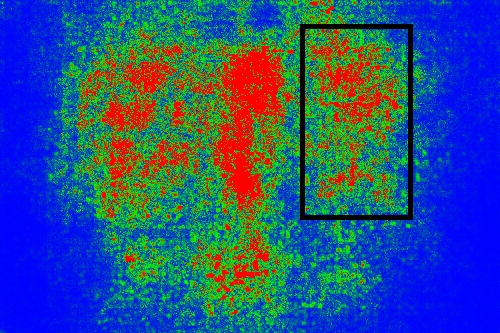}}
\end{minipage}
\hfill
\begin{minipage}[t]{0.24\linewidth}
  \centering
  \centerline{\includegraphics[height=0.95\linewidth,width=0.95\linewidth]{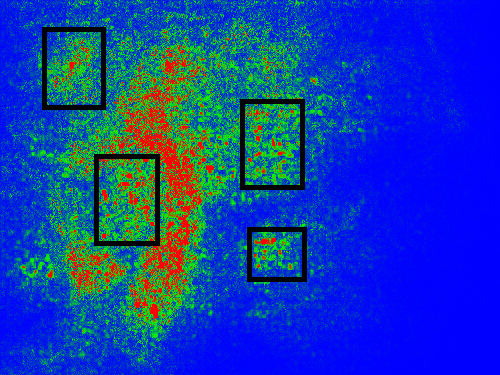}}
\end{minipage}
\hfill
\begin{minipage}[t]{0.24\linewidth}
  \centering
  \centerline{\includegraphics[height=0.95\linewidth,width=0.95\linewidth]{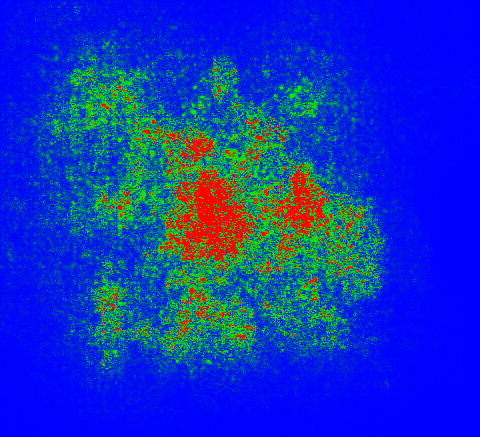}}
\end{minipage}
\hfill
\begin{minipage}[t]{0.24\linewidth}
  \centering
  \centerline{\includegraphics[height=0.95\linewidth,width=0.95\linewidth]{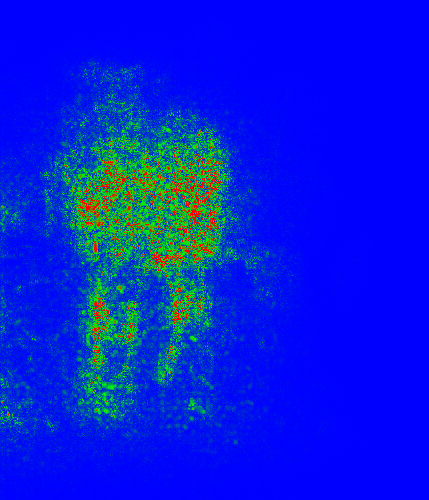}}
\end{minipage}
\vfill
\vspace{4pt}
\begin{minipage}[t]{0.24\linewidth}
  \centering
  \centerline{\includegraphics[height=0.95\linewidth,width=0.95\linewidth]{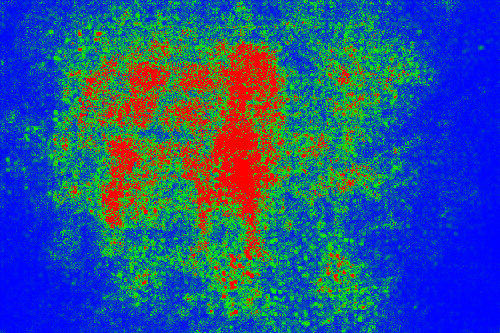}}
\end{minipage}
\hfill
\begin{minipage}[t]{0.24\linewidth}
  \centering
  \centerline{\includegraphics[height=0.95\linewidth,width=0.95\linewidth]{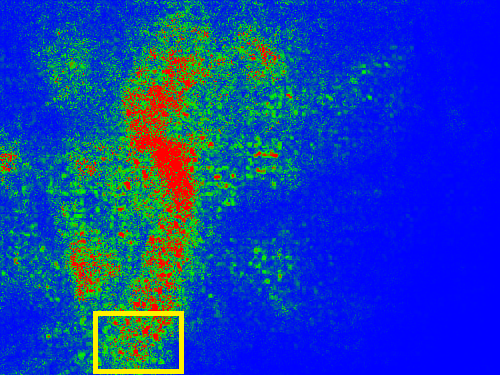}}
\end{minipage}
\hfill
\begin{minipage}[t]{0.24\linewidth}
  \centering
  \centerline{\includegraphics[height=0.95\linewidth,width=0.95\linewidth]{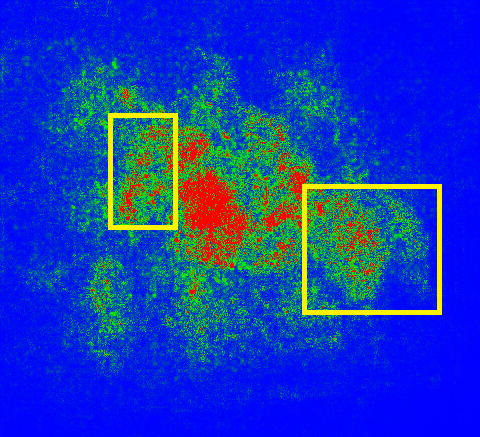}}
\end{minipage}
\hfill
\begin{minipage}[t]{0.24\linewidth}
  \centering
  \centerline{\includegraphics[height=0.95\linewidth,width=0.95\linewidth]{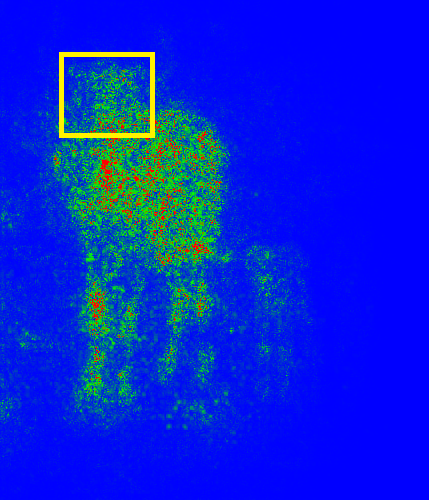}}
\end{minipage}
\end{center}
\caption{ First row: raw images for semantic segmentation. Next two rows are heapmaps for the red cross pixels in the first row's images respectively. Second row is obtained from deeplab largeFOV \cite{chen2016deeplab}, the third row is got from ICNN. Noises distracting attention from objects are framed in black, and valuable information relevant with objects are framed in yellow. Best viewed in color.}
\label{fig:Figure5}
\end{figure}

Figure~\ref{fig:Figure5} shows that irregular kernel can help filter out distractions which may drive attentions away from interested fields. In the first column, conventional convolutional network with regular kernels inevitably augments responses of the undesired ladder fields which change severely in the spatial space. But they are effectively filtered by ICNN in comparison. Figure~\ref{fig:Figure5} also shows that irregular kernels can help take more global information into consideration. As shown in the third column, in order to label the red-cross pixel on the neck, ICNN would further combine the head's information in addition to other parts of the body, which is severely ignored by conventional convolutional network. And with a little more careful observation, we can find that ICNN gets more accurate responses from the hind leg, tail and even stomach. Similar situations happen in the second and forth columns.

\section{Conclusion and Future Work}

The aim of ICNN is to establish the shape compatibility between the input pattern and the kernel. We add a shape attribute for the convolutional kernel and use bilinear interpolation to make it differentiable for end-to-end training. This modification can be smoothly integrated into recent convolutional networks without any extra sub networks. We evaluate our approach in the task of semantic segmentation which has strict requirements for input shape details, and achieve impressive progress. Irregular kernel is a new insight which needs more contributions. First, in practice, considering computation cost, we simplify our approach to one single shared shape for all kernels in a layer. Since different kernels should extract different patterns respectively, efforts should be paid for the full-version ICNN, in which computation acceleration is of great importance. Second, the strategy for interpolation directly determines the update process of the position attribute ${P}$. Detailed information always takes root in closer adjacent pixels while global information usually disperses in entire feature space. We hope more efforts could be devoted to ICNN and to make further progress in deep networks.

{\small
\bibliographystyle{ieee}
\bibliography{egbib}
}

\end{document}